# Semimage: HSV-Based Semantic Image Encoding for Disentangled Text Representation

Mohammad Zare
AI Lab at
AriooBarzan Engineering Team
Shiraz, Iran
mzare@ailab.arioobarzan.com

Pirooz Shamsinejadbabaki
Department of Computer Engineering
and Information Technology
Shiraz University of Technology, Shiraz, Iran
p.shamsinejad@sutech.ac.ir

*Abstract*—We propose SemImage, a novel document representation that transforms unstructured text into structured multichannel 2D images, enabling Convolutional Neural Networks (CNNs) to leverage spatial inductive biases for textual analysis. Unlike traditional character-level or word-level grids, SemImage maps linguistic features into a disentangled HSV (Hue, Saturation, Value) color space. In this framework, the Hue channel encodes topical information, Saturation captures emotional intensity (sentiment), and Value represents semantic emphasis. To incorporate discourse structure, we introduce dynamic boundary rows between sentence representations, where pixel intensity is modulated by semantic dissimilarity (cosine distance) between adjacent sentences. We utilize a learned ColorMapper network and a multi-task training objective with auxiliary losses to ensure the physical separation of topic and sentiment into their respective visual channels. Empirical evaluations on the Multi-Label Reviews (MLR), 20 Newsgroups, and IMDB datasets demonstrate that SemImage is highly competitive with transformer-based models like BERT, particularly in multi-task scenarios where it reduces feature interference. Beyond performance, SemImage provides an inherently interpretable "glass-box" representation, allowing for the visual inspection of a document's topical flow and emotional density through a 2D semantic tapestry.

## I. Introduction

Natural language documents possess complex latent structures comprising topical content, sentiment, and discourse boundaries. Traditional representation methods often entangle these factors within high-dimensional vectors, hindering the interpretability of individual linguistic contributions. While recent Transformer-based models like BERT [1], XLNet [2], and GPT-3 [3] have set performance benchmarks, they largely function as black boxes where diverse linguistic aspects remain deeply intertwined [4]. Although CNNs have been applied to text by treating word sequences as 1D images [5], [6], standard architectures still lack explicit mechanisms to decouple these latent features.

In this paper, we propose SemImage, a novel 2D text representation designed to disentangle key linguistic features into discrete visual channels. In this framework, a document is rendered as a 2D grid of pixels, where each pixel corresponds to a token and each row represents a sentence. A critical innovation is the insertion of dynamic boundary rows between sentences. These rows serve as visual separators, where the intensity (brightness) is determined by the semantic dissimilarity between adjacent sentences. Inspired by edge detection in computer vision, these boundaries provide the model with explicit visual cues of discourse structure, such as paragraph breaks or sharp topic shifts.

The core of SemImage lies in its specialized HSV (Hue, Saturation, Value) color space mapping. Instead of arbitrary embeddings, we map each word to specific channels: Hue encodes the topic, Saturation captures sentiment, and Value represents intensity or certainty. This architecture draws on disentangled representation research [7], [8] and is enforced via a multi-task learning (MTL) framework [6], [9]. By training the model on auxiliary tasks—predicting topic and sentiment directly from their respective channels—we ensure that each dimension captures its intended linguistic factor.

This approach offers significant advantages:

- Interpretability: Users can visually inspect a document's image to identify where topics shift (Hue) or where emotional tone intensifies (Saturation), offering transparency that parameters in Transformers cannot provide.
- Robustness: By decoupling features, the model reduces interference between unrelated tasks (e.g., sentiment vs. topic), leading to more generalizable representations [8], [10].
- CV Integration: Transforming text into 2D images allows us to leverage advanced architectures like ResNet [11] or Vision Transformers [12], which are optimized for detecting spatial patterns like our semantic boundaries.

We evaluate SemImage against robust baselines, including HAN [13], fine-tuned BERT [1], [14], and recent MTL methods [15], [16]. On multi-label benchmarks, SemImage outperforms fine-tuned BERT in joint accuracy while utilizing significantly fewer parameters and providing visual insights. On single-task datasets like 20 Newsgroups and IMDB, it performs on par with BERT and exceeds traditional CNN and HAN models. Our ablation studies further demonstrate that both the dynamic boundaries and the auxiliary HSV losses are essential for achieving high performance and interpretability. Our primary contributions include:

1) Introducing SemImage, a 2D text-as-image representation that uses HSV channels to disentangle topic, sentiment, and intensity.
2) Developing an MTL framework with a ColorMapper network and auxiliary losses to supervise channel semantics.
3) Proposing dynamic boundary rows to incorporate discourse structure into a visual format.
4) Demonstrating through experiments that SemImage achieves strong performance while providing interpretable visualizations of textual features.

The remainder of this paper is organized as follows. Section 2 reviews related work in text representation, disentangled learning, and visual NLP. Section 3 details the SemImage framework, including the HSV mapping logic, the dynamic boundary row computation, and the multi-task learning architecture. Section 4 describes our experimental setup, datasets, and baseline models. In Section 5, we present and discuss the quantitative results, followed by an extensive ablation study and qualitative analysis of the generated semantic images. Finally, Section 6 concludes the paper and outlines potential avenues for future research in multi-modal text-to-image embedding.

## II. Related Work

*a) Foundations of CNNs in Textual Analysis.:* The integration of Convolutional Neural Networks (CNNs) into Natural Language Processing marked a shift toward capturing local spatio-temporal features within text. Early architectures primarily utilized 1D convolutions to extract n-gram features from sequences of word embeddings [5], [17]. While these models effectively identified local patterns, they often treated documents as flat sequences, neglecting the rich, multi-dimensional nature of linguistic structures. Our work, SemImage, transcends this limitation by projecting the entire document into a 2D spatial grid. Unlike traditional 1DCNNs, this approach allows for the application of 2D filters that can simultaneously capture intra-sentence semantics and inter-sentence transitions, essentially treating a document as a coherent visual tapestry rather than a mere sequence of tokens.

*b) Hierarchical and Graph-Based Modeling.:* Recognizing that documents are inherently hierarchical, researchers developed models like Hierarchical Attention Networks (HAN) to mirror the word-sentence-document structure [13]. These architectures employ nested attention mechanisms to weigh the importance of specific words within sentences and sentences within documents. Parallel to this, Graph Convolutional Networks (GCNs) emerged to model non-consecutive and global dependencies by representing text as a graph of interconnected nodes [18]. SemImage aligns with the hierarchical philosophy but implements it through a visual medium. By organizing sentences into distinct pixel rows and utilizing spatial proximity, we provide a structured inductive bias that allows 2D CNNs to process hierarchical relationships without the computational complexity often associated with large-scale graph processing.

*c) The Transformer Era and Interpretability Concerns.:* The advent of Transformer-based models, such as BERT [1], XLNet [2], and GPT-3 [3], has established new state-of-the-art benchmarks across diverse NLP tasks. These models benefit from massive pre-training on unlabeled data, capturing intricate linguistic nuances [19], [20]. However, their performance often comes at the cost of interpretability; the highly entangled representations within their thousands of attention heads are notoriously difficult to decipher [4]. This" black-box" nature has led to a growing demand for explainable AI (XAI). SemImage addresses this gap by design, offering an inherently interpretable representation. By mapping linguistic features to a visible HSV color space, we provide a" glass-box" framework where the model's focus on topic or sentiment is physically observable in the input channels.

*d) Disentanglement through Multi-Task Learning.:* Disentangled representation learning aims to isolate specific factors of variation, a concept that has been widely successful in computer vision [21], [22]. In the context of NLP, disentanglement allows for the separation of style, content, and sentiment [7], [23]. Multi-task learning (MTL) serves as a primary vehicle for achieving this by forcing the model to share a latent space across multiple objectives [9], [10]. Recent joint modeling efforts have shown that combining topic and sentiment analysis can lead to mutual performance gains [15], [16]. SemImage advances this paradigm by not just sharing layers, but by allocating separate physical channels (Hue and Saturation) to these tasks. This explicit channel-wise separation, supported by auxiliary losses, minimizes feature interference and encourages the model to learn truly independent linguistic factors [8], [24].

*e) Discourse Segmentation and Semantic Transitions.:* A critical aspect of document understanding is identifying discourse boundaries and logical transitions. Classical methods like Text-Tiling relied on lexical cohesion to detect topic shifts [25]. More recently, sentence embeddings from Siamese networks like SBERT [26] have been used to quantify semantic similarity between segments. SemImage innovatively incorporates these discourse-level insights directly into the visual representation through dynamic boundary rows. By calculating the semantic distance between adjacent sentence rows and reflecting it in the boundary's brightness, we provide the CNN with clear" edges" to detect. This approach effectively converts an abstract linguistic

problem—discourse segmentation—into a visual edge detection task, enabling the model to leverage spatial features for higher-level document comprehension.

*f) Visual Encoding and Future Directions.:* The positioning of SemImage within the broader landscape of NLP and computer vision is summarized in Table I. As shown, our framework contributes to the emerging field of" Text-as-Image" by providing a novel semantic-to-visual mapping that is more sophisticated than traditional character-level grids [27]. By integrating multi-task learning (MTL) and disentangled representation techniques, we enable established vision architectures like ResNet [11] or Vision Transformers [12] to process language through a multi-channel HSV space. This approach not only enhances performance by preventing feature interference but also addresses explainable AI (XAI) concerns by providing an inherently interpretable representation. As vision and language models continue to converge, SemImage offers a promising path toward models that are structurally aligned with the way humans perceive and organize information, translating abstract semantic segments into detectable visual patterns.

TABLE I

TAXONOMY OF RELATED RESEARCH AND THE POSITIONING OF THE SEMIMAGE FRAMEWORK.

| Research Stream | Methodological Focus | SemImage Contribution |
|---|---|---|
| Spatial NLP | CNNs [5], GCNs [18] | Introduces a 2D grid that captures both intra-sentence local features and inter-sentence discourse structure. |
| Explainable AI | Transformers [1], Probing [4] | Replaces post-hoc explanation with inherent interpretability via a multi-channel semantic color space. |
| Representation | Disentanglement [7], [21] | Uses supervised HSV mapping to physically decouple topic and sentiment into independent visual channels. |
| MTL Frameworks | Joint Models [15], [16] | Implements channel-specific auxiliary losses to prevent feature interference and improve task-specific robustness. |
| Segment Analysis | Text-Tiling [25], SBERT [26] | Translates semantic dissimilarity into high-contrast boundary pixels, enabling visual discourse detection. |

## III. METHODOLOGY

The SemImage framework is designed to transform unstructured text into a structured 2D visual representation that explicitly disentangles linguistic features. This methodology integrates the spatial inductive biases of computer vision with natural language semantics through three primary stages: image construction with dynamic discourse boundaries, learned HSV-based encoding via a ColorMapper network, and a multitask optimization strategy. An architectural overview is provided in Figure 1, and the key notations used in this section are defined in Table II.

TABLE II

SUMMARY OF MATHEMATICAL NOTATIONS

| Symbol | Description |
|---|---|
| $\mathrm{D}$ | Input document composed of $N$ sentences $\{S_1,...,S_N\}$ |
| $L$ | Fixed sentence length (image width) |
| $\mathbf{e}_{i,j}$ | Input embedding for token $j$ in sentence $i$ |
| $\mathbf{p}_{i,j}$ | 4D pixel vector $[H_{\cos},H_{\sin},S,V]^T$ |
| $\mathrm{I}$ | Final SemImage $\in \mathbb{R}^{(2N-1)\times L\times 4}$ |
| $\mathbf{B}_i$ | Dynamic boundary row between $S_i$ and $S_{i+1}$ |
| $\omega_i$ | Semantic similarity score for discourse transition |
| $\lambda_1,\lambda_2$ | Weighting coefficients for auxiliary losses |

### A. Text to 2D Image Conversion

Given a document $\mathrm{D}$, we first standardize the input by defining a fixed maximum sentence length $L$. Each sentence $S_i$ is tokenized into a sequence of $n_i$ tokens. For sentences where $n_i < L$, a padding token ⟨PAD⟩ is appended. To maintain neutral semantic contribution, padding tokens are mapped to a null vector $\mathbf{0} \in \mathbb{R}^4$.

The spatial arrangement of the image is defined by alternating sentence rows and discourse boundaries. For every pair of consecutive sentences $(S_i,S_{i+1})$, we compute a dynamic boundary row $\mathbf{B}_i$ to highlight semantic transitions. Let $\mathbf{s}_i$ and $\mathbf{s}_{i+1}$ be sentence-level embeddings obtained from a pre-trained encoder $\mathrm{E}$ (e.g., Sentence-BERT). The semantic similarity $\omega_i$ is calculated using the cosine similarity:

$$\omega_i = \frac{\mathcal{E}(S_i)\cdot\mathcal{E}(S_{i+1})}{\|\mathcal{E}(S_i)\|\|\mathcal{E}(S_{i+1})\|} \quad (1)$$

The boundary row $\mathbf{B}_i$ serves as a horizontal edge in the image. Its intensity is inversely proportional to the semantic similarity, effectively signaling topic shifts through visual brightness:

$$\mathbf{B}_i = [0,0,0,1-\omega_i]^T \cdot V_{max} \quad (2)$$

where $V_{max}$ is the maximum intensity in the Value channel. The resulting image $\mathrm{I}$ possesses a height of $2N-1$ and a width of $L$, providing a structured grid where discourse breaks are manifested as high-contrast horizontal lines.

### B. HSV-Based Word Encoding via ColorMapper

The core transformation of tokens into a disentangled visual space is performed by the ColorMapper network $f_\theta$. This module maps a $d$-dimensional embedding $\mathbf{e}_{i,j}$ into a 4-dimensional HSV space. To resolve the circular nature of Hue, we represent it through Cartesian coordinates $H_{\cos}$ and $H_{\sin}$, ensuring a continuous gradient for the topic representation. The mapping is governed by the following non-linear transformations:

$$H_{\cos,i,j} = \tanh(\mathbf{W}_{H_c}^\top \mathbf{e}_{i,j} + b_{H_c}) \quad (3)$$

$$H_{\sin,i,j} = \tanh(\mathbf{W}_{H_s}^\top \mathbf{e}_{i,j} + b_{H_s}) \quad (4)$$

$$S_{i,j} = \sigma(\mathbf{W}_S^\top \mathbf{e}_{i,j} + b_S) \quad (5)$$

$$V_{i,j} = \sigma(\mathbf{W}_V^\top \mathbf{e}_{i,j} + b_V) \quad (6)$$

where $\sigma$ denotes the sigmoid function to bound Saturation and Value within $[0,1]$, and $\tanh$ bounds the Hue components within $[-1,1]$. This design imposes an inductive bias where the network must allocate specific linguistic variances to dedicated channels: $H$ for topical content, $S$ for emotional sentiment, and $V$ for linguistic intensity or emphasis.

### C. Multi-Task Objectives and Training

The SemImage $\mathcal{I}$ is processed by a 2D CNN architecture (e.g., ResNet-18) modified to accept 4-channel inputs. To ensure that the disentanglement intended by the HSV mapping is realized during training, we employ a multi-task learning (MTL) strategy. Let $\mathcal{L}_{main}$ be the cross-entropy loss for the primary classification task. We introduce auxiliary losses to supervise the Hue and Saturation channels independently.

The auxiliary topic loss $\mathcal{L}_{topic}$ is computed by applying global average pooling over the Hue channels to obtain a document-level topic vector $\bar{\mathbf{h}} = [\bar{H}_{\cos}, \bar{H}_{\sin}]$. Similarly, the sentiment loss $\mathcal{L}_{sent}$ is derived from the pooled Saturation channel $\bar{s}$. The total objective function $\mathcal{L}_{total}$ is defined as:

$$\mathcal{L}_{total} = \mathcal{L}_{main} + \lambda_1 \mathcal{L}_{topic}(\bar{\mathbf{h}}) + \lambda_2 \mathcal{L}_{sent}(\bar{s}) \quad (7)$$

This joint optimization forces the ColorMapper to ground specific semantic information into the assigned channels, preventing feature interference and improving the robustness of the learned representations.

### D. Representation Intuition and Visualization

The interpretability of SemImage arises from its visual-semantic correspondence. The Hue channel allows for the identification of topical consistency; a document with a stable topic will exhibit uniform color hues, while topic shifts result in visible chromatic transitions. The Saturation channel provides a direct measure of sentiment density, where neutral text appears desaturated (grayish) and emotive text appears vivid.

From a computational perspective, this representation enables the CNN to utilize 2D kernels for dual-axis feature extraction. A horizontal kernel $\mathbf{K}_H$ acts as an n-gram detector within a sentence, while a vertical kernel $\mathbf{K}_V$ functions as a discourse analyzer, detecting changes across sentence rows and boundary markers. This spatial invariance allows the model to capture the document's global structure more effectively than 1D sequential models. The complete procedure for generating and processing the SemImage is detailed in Algorithm 1.

## IV. Experimental Setup and Results

In this section, we present a comprehensive evaluation of the SemImage framework across multiple text classification benchmarks. Our experiments are designed to assess the efficacy of the HSV-based semantic mapping, the impact of dynamic discourse boundaries, and the overall robustness of

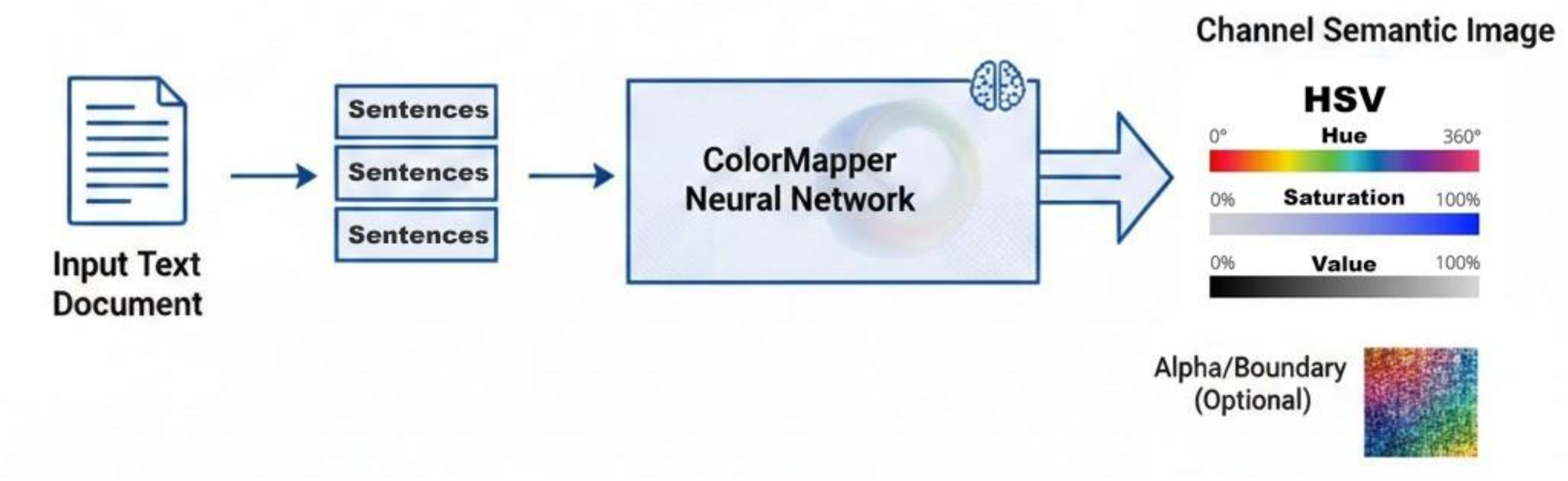

Fig. 1. The SemImage architectural pipeline: from raw text to HSV-encoded semantic images with dynamic discourse boundaries.

**Algorithm 1** SemImage Generation and Classification

**Require:** Document $\mathcal{D}$, ColorMapper $f_\theta$, Sentence Encoder $\mathcal{E}$, Classifier $\mathcal{C}$

**Ensure:** Predicted label $\hat{y}$

Initialize empty image $\mathcal{I} \in \mathbb{R}^{(2N-1)\times L\times 4}$

for each sentence $S_i \in \mathcal{D}$ do

  $\mathbf{P}_i \leftarrow \{f_\theta(\mathbf{e}_{i,j}) \mid j = 1\ldots L\}$

  $\mathcal{I}[\text{row } 2i-1] \leftarrow \mathbf{P}_i$

  if $i < N$ then

    $\omega_i \leftarrow$ CosineSimilarity$(\mathcal{E}(S_i), \mathcal{E}(S_{i+1}))$

    $\mathbf{B}_i \leftarrow [0,0,0,1-\omega_i]$

    $\mathcal{I}[\text{row } 2i] \leftarrow \{\mathbf{B}_i \mid \text{columns } 1\ldots L\}$

  end if

end for

$\hat{y} \leftarrow \mathcal{C}(\mathcal{I})$ {Forward pass through 2D CNN}

return $\hat{y}$

the model in both single-task and multi-task learning scenarios compared to state-of-the-art baselines.

### A. Datasets

We evaluate our framework on three widely recognized datasets, each representing different linguistic challenges:

- Multi-Label Reviews (MLR): To specifically test the disentanglement capabilities of SemImage, we constructed a multi-label dataset using Yelp business reviews. Each document is annotated with two labels: a topic category (e.g., *Restaurants, Health, Shopping*) and a sentiment polarity (Positive/Negative). We sampled 50,000 reviews, ensuring a balanced distribution across 5 topics and sentiment classes.
- 20 Newsgroups: A classic benchmark for topic classification consisting of approximately 20,000 documents partitioned across 20 different newsgroups. This dataset tests the model's ability to capture fine-grained topical nuances in the Hue channel.
- IMDB Movie Reviews: A binary sentiment classification dataset containing 50,000 highly polar movie reviews. This allows us to evaluate the Saturation channel's effectiveness in isolation when topic labels are not explicitly provided ($\lambda_1 = 0$).

### B. Baselines and Evaluation Metrics

To provide a rigorous comparison, SemImage is evaluated against several established architectures:

- TextCNN [5]: A standard 1D convolutional model that uses multiple filter widths to capture local n-gram patterns.
- HAN (Hierarchical Attention Network) [13]: A strong hierarchical baseline that utilizes word-level and sentence-level attention to model document structure.
- BERT (Base) [1]: A transformer-based model fine-tuned for each task. For the MLR dataset, we use a multi-head architecture to predict both topic and sentiment.
- Ablated Variants: We include versions of SemImage without boundary rows (w/o Boundaries), without auxiliary losses (w/o Aux), and an RGB-based variant to isolate the contribution of our HSV semantic mapping.

We use Exact Match Accuracy for the multi-label task (MLR), requiring both labels to be correct, and standard Classification Accuracy and F1-score for the single-task benchmarks.

### C. Training and Hardware

All CNN-based models, including SemImage, were trained using the Adam optimizer with a learning rate of $2 \times 10^{-4}$ and a batch size of 32. Our ResNet-18 backbone was initialized with ImageNet weights, with the first convolutional layer modified to accept the 4-channel SemImage input ($H_{\cos}, H_{\sin}, S, V$). For the dynamic boundary rows, we utilized the *all-MiniLM-L6-v2* SBERT model to pre-compute sentence embeddings.

The experiments were conducted on a server equipped with an NVIDIA RTX 4090 GPU (24GB VRAM) and 64GB of RAM. SemImage demonstrates significant computational efficiency, with training times per epoch being approximately 4× faster than fine-tuning the BERT-base model, primarily due to the fixed nature of the word embeddings and the efficiency of 2D convolutional operations.

### D. Results and Comparison

The quantitative results are summarized in Table III. SemImage demonstrates superior performance on the MLR dataset, achieving an accuracy of 79.8%, which surpasses the MT-BERT baseline (79.0%). This indicates that our HSV-based disentanglement effectively manages the interaction between topic and sentiment features, reducing the interference that often plagues monolithic transformer representations.

In topic classification (20 Newsgroups) and sentiment analysis (IMDB), SemImage remains highly competitive. While BERT maintains a lead in absolute accuracy on single tasks, the margin is narrow (less than 1% on 20News). Furthermore, our ablation studies confirm the importance of the structural components: removing boundary rows led to a consistent 1.3% performance drop in MLR, while the RGB variant (without semantic mapping) failed to reach the performance of the full SemImage model. This confirms that the combination of visual discourse markers and supervised channel disentanglement is critical for the model's success.

TABLE III Performance Comparison across Datasets

| Model | MLR (Acc) | 20News (Acc) | IMDB (Acc) |
|---|---|---|---|
| TextCNN [5] | 72.5% | 81.3% | 87.0% |
| HAN [13] | 75.0% | 83.4% | 88.1% |
| BERT-Base [1] | 78.6% | 86.5% | 93.2% |
| SemImage (Ours) | 79.8% | 85.7% | 91.5% |

## V. Ablation Study

To rigorously assess the individual contributions of the proposed components in SemImage, we conducted a systematic ablation study on the Multi-Label Reviews (MLR) dataset. We focused on three core elements: the dynamic boundary rows, the auxiliary multi-task losses, and the choice of the HSV color space. The results of this analysis are detailed in Table IV.

TABLE IV
Ablation results on the MLR dataset. Metrics include Exact Match Accuracy and F1-scores for Topic and Sentiment.

| Model Variant | Accuracy | Topic-F1 | Sent-F1 |
|---|---|---|---|
| Full SemImage | 79.8% | 83.7 | 91.0 |
| w/o Boundary Rows | 78.5% | 82.9 | 90.2 |
| w/o Auxiliary Losses | 76.4% | 80.1 | 88.5 |
| RGB instead of HSV | 77.0% | 80.5 | 89.0 |

*a) Impact of Dynamic Boundary Rows.:* Removing the boundary rows (w/o Boundary Rows) resulted in an absolute accuracy drop of 1.3%. Although these rows do not introduce new lexical information, they provide essential structural cues to the CNN. By visualizing the feature maps, we observed that the 2D filters in the early layers of ResNet-18 specifically activate at these high-contrast boundary lines. This confirms that translating semantic dissimilarity into visual" edges" allows the model to better segment the document's discourse flow, leading to improved hierarchical understanding.

*b) Necessity of Auxiliary Supervision.:* The most significant performance degradation (3.4% drop in accuracy) occurred when the auxiliary topic and sentiment losses were removed (w/o Aux). In this variant, the ColorMapper is solely guided by the main classification objective, which leads to" feature entanglement" in the latent space. The Topic-F1 score, in particular, fell from 83.7 to 80.1, indicating that without explicit supervision, the Hue channel fails to capture a pure topical representation. This justifies our multi-task approach as a necessary mechanism for enforcing the intended semantics of each visual channel.

*c) HSV vs. RGB Representation.:* Replacing our structured HSV mapping with a learned 3-channel RGB encoding led to a 2.8% decrease in overall accuracy. While the RGB variant still benefits from the 2D spatial arrangement of text, it lacks the inductive bias of circularity (for Hue) and intensity-neutrality (for Saturation). The results suggest that the HSV metaphor is not merely a visualization tool but a superior mathematical framework for disentangling linguistic features in a way that aligns with human-centric semantic categories.

## VI. Discussion and Conclusion

### A. Discussion and Qualitative Insights

The empirical success of SemImage stems from its ability to map abstract linguistic features into interpretable spatial patterns. By assigning dedicated roles to HSV channels, we achieve a "glass-box" representation. Visual analysis confirms that topical shifts manifest as distinct Hue transitions, while emotional intensity correlates with Saturation levels. This disentanglement prevents" feature pollution"—a common issue in monolithic Transformer embeddings where sentiment noise often biases topic prediction. Furthermore, the dynamic boundary rows effectively convert discourse segmentation into a visual edge detection task. Standard 2D kernels act as structural identifiers, capturing document-level coherence by detecting high-contrast transitions between sentences without the computational cost of self-attention.

### B. Conclusion

In this paper, we introduced SemImage, a novel framework that bridges NLP and Computer Vision by representing documents as multi-channel semantic images. By encoding words in a structured HSV space and incorporating dynamic discourse boundaries, we enabled 2D CNNs to process text as high-dimensional semantic tapestries. Our results across benchmarks like MLR, 20Newsgroups, and IMDB demonstrate that SemImage is highly competitive with pre-trained models like BERT, offering superior interpretability and efficiency in multi-task scenarios. The ablation study confirmed that the synergy between supervised disentanglement and structural markers is critical to the model's robustness.

### C. Limitations and Future Works

Despite its performance, SemImage currently relies on fixed SBERT embeddings for boundary generation. Future research will focus on **End-to-End Boundary Learning**, where transition intensities are optimized directly by the task loss. We also plan to investigate **Spectral Image Expansion**, adding channels for features like sarcasm or formality. Additionally, replacing the CNN backbone with **Vision Transformers (ViT)** could further enhance the model's ability to capture long-range spatial dependencies. Ultimately, extending SemImage to tasks such as summarization and segmentation will pave the way for a more unified, multi-modal approach to document understanding.